\documentclass[final,12pt]{clear2025} 
\usepackage{times} 

\title{Interpretable Neural Causal Models with TRAM-DAGs}

\usepackage{hyperref}  
\usepackage{xcolor}    
\usepackage{booktabs}

\usepackage{dsfont}
\usepackage{xcolor} 
\usepackage[capitalise]{cleveref}

\newcommand{\pSL}{F_{\SL}}

 \DeclareMathOperator{\SL}{SL}

 \DeclareMathOperator{\odds}{odds}

\newcommand{\revision}[1]{{#1}}
\newcommand{\pa}[1]{\text{pa}(#1)}
\newcommand{\dd}[1]{\text{do}(#1)}

\clearauthor{%
 \Name{Beate Sick} \Email{beate.sick@uzh.ch}\\
 \addr UZH, Zurich, Switzerland \\
 \addr ZHAW, Winterthur, Switzerland
 \AND
    \Name{Oliver Dürr} \Email{oliver.duerr@htwg-konstanz.de}\\
    \addr HTWG Konstanz, Germany\\
    \addr TIDIT.ch, Switzerland
}

\begin{document}
\maketitle
\begin{abstract}
The ultimate goal of most scientific studies is to understand the underlying causal mechanism between the involved variables. Structural causal models (SCMs) are widely used to represent such causal mechanisms. Given an SCM, causal queries on all three levels of Pearl's causal hierarchy can be answered: $\mathcal{L}_1$ observational, $\mathcal{L}_2$ interventional, and $\mathcal{L}_3$ counterfactual.
An essential aspect of modeling the SCM is to model the dependency of each variable on its causal parents. Traditionally this is done by parametric statistical models, such as linear or logistic regression models.
This allows to handle all kinds of data types and fit interpretable models but bears the risk of introducing a bias due to the assumed rigid functional form. 
More recently neural causal models came up using neural networks (NNs) to model the causal relationships, allowing the estimation of nearly any underlying functional form without bias. However, current neural causal models are generally restricted to continuous variables and do not yield an interpretable form of the causal relationships. 
Transformation models range from simple statistical regressions to complex networks and can handle continuous, ordinal, and binary data. 
Here, we propose to use \revision{potentially deep} TRAMs to model the functional relationships in SCMs allowing us to
bridge the gap between interpretability and flexibility in causal modeling.
We call this method TRAM-DAG and assume currently that the underlying directed acyclic graph (DAG) is known. 
For the fully observed case, we benchmark TRAM-DAGs against state-of-the-art statistical and NN-based causal models. We show that TRAM-DAGs are interpretable but also achieve equal or superior performance in queries ranging from $\mathcal{L}_1$ to $\mathcal{L}_3$ in the causal hierarchy. 
For the continuous case, TRAM-DAGs allow for counterfactual queries for three common causal structures, including unobserved confounding.
\end{abstract}

%

%

\section{Introduction}
Causal understanding is the ultimate goal in science and also essential in applications such as healthcare, economics, and policy-making because it allows to design effective interventions and make well-founded decisions.
Structural Causal Models (SCMs) have become an established method for a mathematical representation of causal models. 
An SCM allows to tackle tasks on all three levels of Pearl’s causal hierarchy: fitting observational distributions (Level 1), estimating interventional distributions (Level 2), and answering counterfactual queries (Level 3) \citep{pearl2000models}. 
One line of research in causal modeling is to estimate the directed acyclic graph (DAG), capturing the existence and directions of these mutual causal relationships as far as possible from observational data and asses if there are unobserved confounders. Another line of research starts from the DAG and focuses on estimating the functional form of the causal relationships of each variable on its causal parents. 
In this paper, we focus on this second line of causal modeling research and introduce TRAM-DAGs (see \cref{fig:trams}). 
The approaches for modeling the causal relationships can often be assigned to one of two choices:
1) A statistical approach that offers transparent parametric models for capturing causal relationships between variables of all data types but is prone to bias due to restrictive assumptions about functional forms (\cite{peters2017elements}). 
2) Neural network (NN) based causal models 
allowing for an unbiased estimation of complex relationships but suffering from their restriction to continuous data and their black box character. 

To the best of our knowledge, our suggested TRAM-DAG is the first interpretable neural causal model that: a) can handle continuous, ordinal, binary or mixed data types and b) comprises classical statistical and NN-based approaches allowing to model the causal relationships of the SCM with interpretable or fully flexible model-parts within the same framework. 
We demonstrate that TRAM-DAGs achieve at least state-of-the-art performances in answering causal queries across the three levels of Pearl's causal hierarchy while retaining the interpretability required for understanding causal relationships. 

\section{Existing approaches for estimating causal relationships in SCMs}
Estimating the functional relationships in SCMs can be broadly divided into statistical and NN-based approaches. 

\subsection{Causal models based on neural networks}
There is a growing body of literature on estimating causal relationships using NNs, mostly generative neural network models, see  \cite{poinsotLearningStructuralCausal2024a} for a recent comprehensive review.
Theoretical results on the identifiability of neural causal models with continuous variables are discussed in \cite{xia2023neural} for all three levels of causal hierarchy: fitting the observational data ($\mathcal{L}_1$), estimating intervention effects and interventional distributions ($\mathcal{L}_2$) and answering counterfactual questions ($\mathcal{L}_3$).
These results are supplemented by numerical experiments using simple feed-forward NNs. 
Other methods go beyond simple feed-forward NNs. E.g. \cite{sanchez-martinVACADesigningVariational2022} introduced  Variational Graph Autoencoders (VACA).
In VACA, the encoder graph network is not allowed to have hidden layers to allow for $\mathcal{L}_3$ identification,
which limits the expressiveness of the approach. 
\revision{Handling \(\mathcal{L}_2\) queries is also possible with sum-product network, see \cite{poonia2024chi} or circuit models \cite{wang2023compositional}}.
A particularly interesting class of causal models \revision{capable of \(\mathcal{L}_3\) queries} are models with a bijective generation mechanism (BGM) as described by \cite{nasr2023counterfactual}. Their work demonstrated that models in this class are identifiable in the fully observed and two other cases \revision{with unobserved confounders, i.e.~an instrumental variable or a backdoor setting}. Specifically, a BGM model trained to fit continuous observational data at \(\mathcal{L}_1\) can also predict \(\mathcal{L}_2\) and \(\mathcal{L}_3\) queries.
However, to be bijective BGMs are restricted to continuous variables (see \cref{sec:sample_obs} for a discussion). 
Prominent members of that class of BGMs are normalizing flows (NFs). NFs rely on a single or a series of simple, invertible transformations to map variables to a simpler latent
distribution - hence NF and TRAMs rely on the same idea (see \cref{sec:trams}).
Initially, NFs have been proposed for causal estimation by \cite{khemakhemCausalAutoregressiveFlows2021}, who introduced CAREFL that relies on chaining many simple transformations. 
Other recent NF-based methods achieve flexible transformations without chaining
by directly modeling monotonic transformation. E.g., \cite{balgi2024deep} uses unconstrained monotonic NNs. However, all current NF-based methods 
suffer from their black box character and are generally restricted to continuous data. 

\subsection{Causal models based on statistical models}

Causal modeling dates back to the 1920s when path diagrams were introduced to describe causal relationships by \cite{wright1920relative}. 
 To describe the functional relationships in these causal models, structural equations like discussed in \cref{fig:dag_obs} were formally developed in the 1970s by 
\cite{joreskog1970general}. Pearl introduced 1995 the do-calculus based on DAGs for answering causal queries \cite{pearl1995causal}.
While often linear regression models were used in the past, modern approaches can model more complex structures and use e.g.\ generalized linear models (GLMs) to set up the structural equations. 
Classical statistical models are usually not over-parametrized, and  their interpretable parameters can be consistently estimated.
A drawback of statistical modes are their limited flexibility, which can lead to suboptimal estimates of observational and interventional distributions. 
An overview of causal inference in statistics can be found in \cite{pearl2009causal} and a detailed discussion of complete identification methods for the causal hierarchy in \cite{shpitser2008complete}. 
In our study, we are in a quite easy setting because we assume a fully observed DAG, and utilize well-characterized transformation models to estimate the causal relationships. Hence, \(\mathcal{L}_2\) queries can be solved by the do-calculus of Pearl when the observational data is accurately fitted \cite{pearl1995causal}.  

\section{Background} 
\label{sec:bg}
\begin{figure}[h!]
    \centering
    \begin{minipage}[b]{0.81\linewidth}
        \centering
        \includegraphics[width=\linewidth]{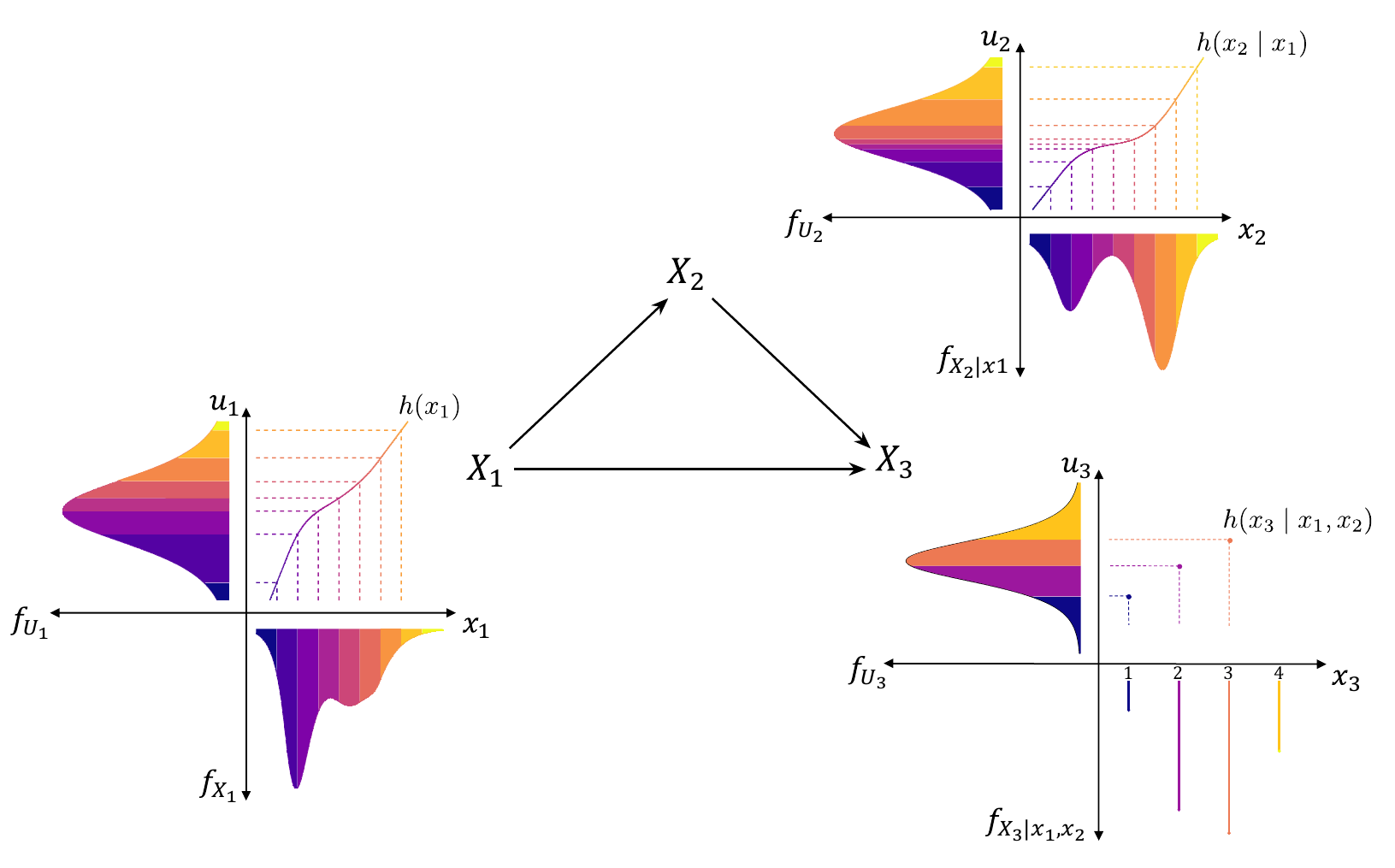}
        \caption{TRAM-DAG model: Each variable in the DAG is modeled by a TRAM. The \revision{TRAMs of the} continuous variables $X_1$ and $X_2$ have a continuous $h$, and \revision{the TRAM of the discrete} ordinal variable $X_3$ has a discrete $h$. 
    For variables with parents, the conditional transformation function $h(X_i|\pa{X_i})$ and outcome distribution $f_{X_i|\pa{X_i}}$ depend on the values of the parents. }
        \label{fig:trams}   
         \end{minipage}
 \end{figure}

%
We briefly introduce the necessary background needed for the proposed TRAM-DAG method (see \cref{fig:trams}) for causal modeling. 

\subsection{Background on DAGs and SCMs for TRAM-DAGs}
\label{sec:bg_dags}

\begin{figure}[h!]
    \centering
    \begin{minipage}[b]{0.49\linewidth}
        \centering
        \includegraphics[width=\linewidth]{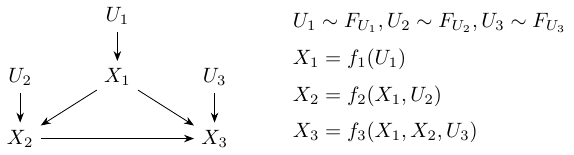}
        \label{fig:dag_obs}
    \end{minipage}
    \hfill
    \begin{minipage}[b]{0.49\linewidth}
        \centering
        \includegraphics[width=\linewidth]{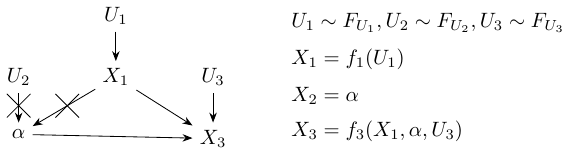}
        \label{fig:dag_do}
    \end{minipage}
    
    \caption{\textbf{Left:} DAG and corresponding SCM skeleton for three observed variables $X_1, X_2, X_3$ and unobserved noise $U_1, U_2, U_3$. \textbf{Right:} The post-interventional DAG and SCM skeleton when performing a $\dd{X_2=\alpha}$ intervention.}
    \label{fig:dags}
\end{figure}

In this study, we assume that the underlying causal structure, given by a directed acyclic graph (DAG) (see \cref{fig:dags}), is known. We focus on the case where all variables are observed. However, we would like to emphasize that for the continuous case, TRAM-DAGs are in the class of bijective generation models (BGM) (see \cref{proof}) and therefore applicable beyond the full observed case \cite{nasr2023counterfactual}. 
For the ease of discussion, we restrict to causal models
where we have $d$ mutual independent noise variables $U_i$ 
meaning
that we have no unobserved confounders.
Although noise variables are typically omitted from DAGs for clarity, we include them in the  DAG shown in \cref{fig:dags}.  

Taken together, the collection of of $d$ (structural) assignments, 
\begin{equation}
X_i := f_i(\pa{X_i}, U_i), \quad \text{for } i = 1, \ldots, d.
\label{eq:SEM}
\end{equation}
and the specification of the $d$ mutual independent noise distributions $F_{U_i}$ define a structural causal model (SCM)  for the involved variables $X_1,...,X_d$.
Please note that given the observed data, the form of the functions $f_i$ in an SCM would change if another noise distribution $F_{U_i}$ is assumed. 

When performing a deterministic do-intervention in a causal model, for example, $\dd{X=\alpha}$, the intervened variable is forced to take the value of the do-intervention. 
Consequently, no other variables influence the intervened variable, resulting in a post-intervention DAG where all directed edges pointing to the intervened variable are removed (see right panel of \cref{fig:dags}). The post-interventional SCM is updated only for the intervened variable, which now takes on the fixed value imposed by the intervention.

\subsection{Background on transformation models as needed for TRAM-DAGs} 
\label{sec:trams} 
 We use (deep) transformation models (TRAMs), \revision{which have so far only been used in non-causal regression tasks} \citep{hothorn2014conditional, sick2021deep, kook2020ordinal, baumann2021deep}, \revision{for causal modeling by using the causal parents of a variable as predictors (\cref{fig:trams})}.
%
\revision{We can directly use the} core idea of TRAMs \citep{hothorn2014conditional} to  construct the conditional distribution of a variable $X_i$ in a causal model, whether continuous, ordinal, or binary, on its parents as follows:
A strictly monotone increasing conditional transformation function  $h(x_i\mid \pa{x_i})$  is fitted that maps the unspecified conditional outcome distribution \( F_{X_i \mid \pa{X_i}} \) to a fixed continuous latent distribution $F_u$  with a log-concave and continuous density $f_u$ (see \cref{fig:trams}).  
This approach allows to model the conditional outcome distribution $P(X_i\leq x_i|\pa{X_i})=F_{X_i|\pa{X_i}}(x_i)$,   as 
\begin{align} \label{eq:tram}
    F_{X_i|\pa{X_i}}(x_i) = F_U(h(x_i|\pa{x_i})).
\end{align}

\paragraph{Structure of the transformation function $h$:} The transformation function $h=h_I + h_S $ for a target $X_i$ consists of an intercept $h_I$ that can depend on the parents $\pa{X_i}$ and potentially a shift term $h_S$ that can potentially consist of a sum of several linear and complex shift term that depend on the parents. 
Depending on the structure of $h$, the TRAM can be interpretable or be complex allowing for a high flexibility (see \cref{sec:interpretation}). \paragraph{Choice of the latent distribution $F_U$:} The interpretation scale of the shift terms in $h$ depends on the choice of $F_u$, while $F_u$ does not impact the ability to accurately estimate conditional outcome distributions \citep{hothorn2014conditional}. In this study, we always use the standard logistic distribution as $F_u$ since it allows us to interpret the shift terms in $h$ as log-odds-ratios (see \cref{sec:A_interpretability}).

\paragraph{Intercept function for discrete ordinal or binary variables:} 
For an ordinal variable $X_i$ with levels $1, 2,...K$, we use a monotone increasing discrete function to model the intercept function $h_0$ (see e.g.\ $X_3$ in \cref{fig:trams}).  The discrete intercept function, which potentially depends on the parents, is given by:
\begin{equation}
\small
\label{eq:step_h}
    h_I(X_i=k|\pa{X_i}) = \vartheta_k(\pa{X_i}) 
\end{equation} 
where $\vartheta_k$ is a strictly monotone increasing sequence for $k=1,...K$. The probability for a class level $k  \in \{2,...,K-1\}$ is given by the area under the latent density over the interval $[\vartheta_{k-1}, \vartheta_k]$, for $k=1$ the interval is $[-\infty, \vartheta_1]$, for $k=K$ it is $[\vartheta_{K-1},\infty]$.  
A binary outcome can be seen as a special case where $h$ only consists of one cut-point $\vartheta$, cutting the area under latent density in two parts where the lower part represents $P(X_i=0)=F_U(h(x_i=0)$.

\paragraph{Intercept function for continuous variables:} 
For continuous variables, we use Bernstein polynomials to model the continuous intercept function, which potentially depends on the parents. 
\begin{equation}
\small
\label{eq:bernstein_h}
    h_I(x_i|\pa{X_i}) = \frac{1}{M+1} \sum_{k=0}^M \vartheta_k(\pa{X_i})  \mathrm{Be}_{k, M}(x_i), 
\end{equation}
where $\vartheta_k, k=0,...,M$ are strictly monotone increasing coefficients of the Bernstein polynomial to ensure a strictly monotone increasing $h$ and $\mathrm{Be}_{k, M}(x_i)$ denotes the density of a Beta distribution with parameters $k+1$ and $M-k+1$. 
We choose Bernstein polynomials because they can easily be restricted to be strictly monotonically increasing 
and provide theoretical guarantees for approximating any conditional continuous distribution arbitrarily well as long as the order $M$ is sufficiently large \cite{hothorn2014conditional}. 
For such a bijective $h$ we can directly formulate $X_i=f_i(\pa{X_i}, U_i)=h^{-1}(U_i|\pa{X_i})$. 

\subsubsection{Flexible and Interpretable Deep TRAMs: CI, SI-CS, SI-LS  }\label{sec:interpretation}
To model \textbf{fully flexible} function $f_i$ for a variable $X_i$ in an SCM we allow the parameters $\vartheta_k$ in the intercept of $h$ \revision{(discrete or continuous)} to change with the value of the parents of $X_i$ (see \cref{eq:step_h}, \cref{eq:bernstein_h}). We call this a complex intercept (CI) model,  
\revision{which provides maximal flexibility and can approximate any conditional outcome distribution arbitrarily well,  
as shown in \cite{hothorn2014conditional}, where CI models are referred to as response-varying effect models.}


To model a \textbf{causally interpretable} effect for each parent $X_j \in \pa{X_i}$ on $X_i$, we design the transformation function as $h(x_i|\pa{x_i})=h_I(x_i) + \sum_j s(x_j)$ with a simple intercept (SI) $h_I$ that does not depend on the parents, and \revision{additive} interpretable shift terms $s(x_j)$ depending on the values $x_j$ taken by the parents $X_j \in pa(X_i)$. 
A shift term $s(x_j)$ is either a linear shift (LS) $\text{LS}_{x_j}=\beta_j x_j$ or a complex shift (CS) $\text{CS}_{x_j}=\gamma(x_j)$.
In this study, all shift parameters in the TRAM-DAG can be interpreted as log-odds-ratios since we use in all experiments the standard logistic distribution as $F_U$. 
\revision{While being the least flexible, the linear shift terms in the transformation $h$ are the most interpretable. The parameter $\beta_j$ can be causally interpreted as the log-odds ratio. Hence, $\exp(\beta_j)$ represents the factor by which the odds, $\odds(X_i \leq x) = \frac{P(X_i \leq x)}{1-P(X_i \leq x)}$, change when intervening on the parent $X_j\in \pa{X_i}$ by increasing it by one unit (see \cref{sec:app_trams}).   
Importantly, in causal models, we do not require that all other parents $X_{j'} \in \pa{X_i}$ with $j' \neq j$ stay constant; they may also change upon the intervention on $X_j$. \cref{sec:a_interpretation_mixed} provides an illustrative experiment showing that the causal parameter $\beta_j$, that was estimated on observational data, can be used to correctly predict the interventional effect of a parent $X_j$ on the target $X_i$, by comparing $e^{\beta_j}$ to the change of the $\odds(X_i \leq x)$ when the intervention is actually performed in the DGP by increasing the parent $X_j$ by one unit.
For the more flexible complex shift terms $\text{CS}_{x_j}=\gamma(x_j)$, the change in the odds when increasing $X_j$ by one unit  
can be expressed as $\exp(\gamma(x_j+1) - \gamma(x_j))$. While the causal effect of a CS cannot be summarized by a single coefficient anymore, it can be interpreted by plotting $\gamma(x_j)$ against $x_j$ (see e.g.~\cref{fig:DGP_cont}).
}

\revision{
To decide between a TRAM with full flexibility or interpretable effects, we follow the top-down approach as described in \cite{Hothorn_topdown} aiming for maximal interpretability without sacrificing too much predictive performance as measured by the likelihood on an independent test set.
}

\section{TRAM-DAGs}
\label{sec:tram-dag}
\begin{figure}[ht]
    \centering
    \includegraphics[width=0.6\columnwidth]{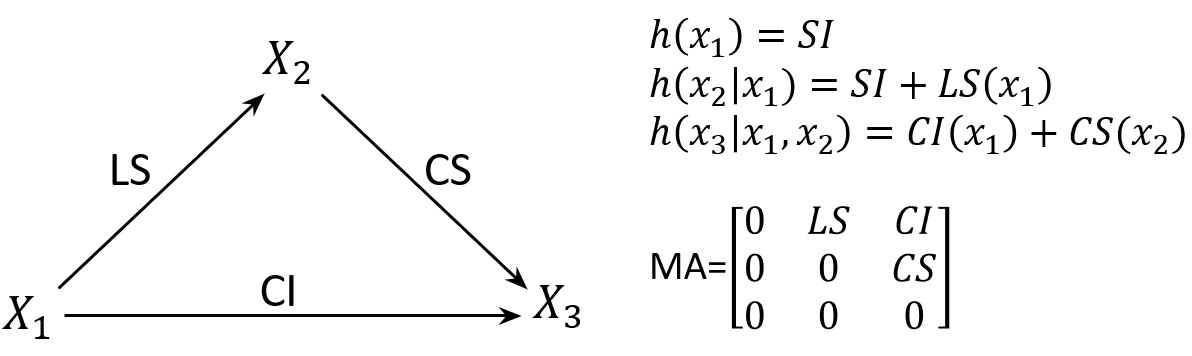} 
    \caption{\textbf{Left}: DAG with meta information on the TRAMs, \textbf{Right:} $h$ structures and the meta-adjacency matrix MA resulting from the DAG and the TRAM structures.}
    \label{fig:MA}
\end{figure}

Here, we describe briefly how to fit TRAM-DAGs (see \cref{fig:trams}) and how to use them for tackling causal queries on all three levels in Pearl's causal hierarchy. The structure of TRAM-DAGs can be described by meta-adjacency matrix MA (see, e.g., \cref{fig:MA}) where the element in the $i$-th row and $j$th column describes the effect type of $X_i$ on $X_j$ which can be either a complex intercept (CI), a complex shift (CS), a linear shift (LS) or no influence at all (0). If the $j$-th column holds no CI-entry, then the intercept is modeled as SI, if all entries are 0, then $X_j$ is a source node with $h(x_j)=SI$. 

\subsection{Architecture and training of TRAM-DAGs}
\label{sec:MAF}
Each (deep) TRAM in the TRAM-DAG can be trained separately on observational data. Details on training a (deep and interpretable) TRAM based on NNs are described in \cite{kook2020ordinal, herzog2023deep, kook2022estimating}.
For convenience, we have constructed a model that consists of a set of customized NNs including masked autoregressive flows (MAFs, see \cite{papamakarios2018maskedautoregressiveflowdensity}) taking as input the meta-adjacency matrix MA (see \cref{fig:MA}) and the observational data. The output is
the components of the $d$ transformation functions, i.e.\ the $\vartheta$-values for the intercept terms (see \cref{eq:step_h}, \cref{eq:bernstein_h}) and the linear and complex shift terms. We train all NNs jointly using the Adam optimizer.

\subsection{Causal queries using TRAM-DAGs}\label{sec:proof}

If all variables are continuous and the TRAMs are flexible enough to accurately fit the observational data, we can show that our fitted TRAM-DAG can solve tasks on all three levels of Pearl's causal hierarchy.
\paragraph{Proposition: Counterfactual Equivalence of continuous TRAM-DAGs}
\ \newline
Consider a 
SCM 
that contains only continuous variables.
%
If  
the continuous \revision{and fully observed} TRAM-DAG model can reproduce the observational distribution $\mathcal{L}_1$, then 
it will also reproduce the same interventional $\mathcal{L}_2$ and counterfactual $\mathcal{L}_3$ queries as  
the SCM.

\begin{proof}\label{proof}
A continuous TRAM-DAG is based on transformation functions modeled by Bernstein polynomials with strictly monotone increasing coefficients. 
This ensures that each transformation is strictly monotone increasing.  
Therefore, continuous TRAM-DAGs fall into the class of BGMs of and the  Lemma B.2 in \cite{nasr2023counterfactual} holds, stating the equivalence.
\end{proof} 
%
Although we focus on the fully observed case in this paper, it is important to note that continuous TRAM-DAGs, are BGMs and so theoretically capable of handling also \revision{some additional} scenarios with unobserved confounders, as demonstrated in Lemma B.3 and B.4 in \cite{nasr2023counterfactual}.
\revision{Please also note that the monotonicity constraint does not effect the expressiveness of the transformation.}
%
In the following, we will show how tasks on $\mathcal{L}_1$ and $\mathcal{L}_2$ are tackled by all kinds of TRAM-DAGs and tasks of $\mathcal{L}_3$ by continuous TRAM-DAGs. 

\textbf{$\mathcal{L}_1$: Sampling from the observational distribution of continuous or mixed TRAM-DAGs}\, 
\label{sec:sample_obs}
To estimate the joint observational distribution, we sample $d$-dimensional observations $(x_1,...,x_d)$ from our fitted TRAM-DAG. 
We first sample values $u_{j}$ from each exogenous distribution $u_{j} \sim F_{U_j}, j=1,...d$. We go along the causal order and start with source nodes. In the case of a discrete variable $X_i$, we increase the sampled $u_i$ to the next value of the discrete transformation function $h\geq u_i$. 
We deduce the corresponding sample $x_{i}$, by the requirement $u_{i}=h(x_{i})$ which in case of continuous variables can be written as $x_{i}=h^{-1}(u_i)$. 
For non-source nodes $X_j$, 
$u_j =h(x_j|\pa{x_j})$ is determined by the sampled values $x_j$ of the parents. To get a sample $x_j$ corresponding to $u_j$, we proceed as before.

\textbf{$\mathcal{L}_2$: Estimating interventional distributions and treatment effects for continuous or mixed TRAM-DAGs}\,
We look at do-interventions where one variable is set to a certain value $\dd{X_i=\alpha}$. This results in a post-interventional graph where all arrows pointing to the intervened variables are deleted (see Figure \ref{fig:dags}). To estimate the joint interventional distribution, we proceed similarly as described in $\mathcal{L}_1$ 
but set $X_i=\alpha$ and go now along the causal order in the post-interventional DAG.

\textbf{$\mathcal{L}_3$: Answering counterfactual queries with continuous TRAM-DAGs}\,
In a counterfactual task, we answer "what if" questions such as: What value would $X_J$ have taken if the variable $X_i$ had taken the values $\alpha$ instead of the observed value $x_i$?
For illustration, look at \cref{fig:trams} and assume $X_j=X_3$ and $X_i=X_2$ and that all variables are continuous.
We answer the counterfactual question in three steps:
\newline
1) Abduction: determine the  noise values $u_i$ that correspond to the observations by $u_i=h(x_i|\pa{X_i})$,
\newline
2) Action: Determine the counterfactual DAG for $x_i^c=\alpha$ that correspond to the post-interventional DAG (see for example \cref{fig:dag_do}),
\newline
3) Prediction: use the counterfactual DAG, the counterfactual value $x_i^c=\alpha$, the sampled noise values and go along the causal order to determine for all descendants $X_j$  the counterfactual value as $x_j=h^{-1}(u_i|\pa{x_j}$ by using the updated $h$ where at least one parent is a descendant of $X_i$ and has, therefore, an updated counterfactual value which likely resulted in a changed $h$. 
Note that for discrete or mixed TRAM-DAGs, counterfactual queries are not possible \revision{which is a fundamental limitation of causal models based on discrete ordinal variables, generated by interval censoring of an underlying continuous latent variable, and not a limitation of our proposed framework} (see \cref{sec:l3_discrete} for details). 

\section{Benchmarking TRAM-DAG against Neural Causal Models}
Here, we benchmark TRAM-DAG with state-of-the-art NN and NF-based causal models. These models focus on estimating observational, interventional, and counterfactual distributions without requiring the interpretability of the fitted SCM. The code to reproduce the experiments can be found at: \url{https://github.com/tensorchiefs/tram-dag}.

\subsection{Observational Distribution $\mathcal{L}_1$}
\label{sec:vaca_l1}
\begin{figure}[ht]
    \centering
    \includegraphics[width=0.8\columnwidth]{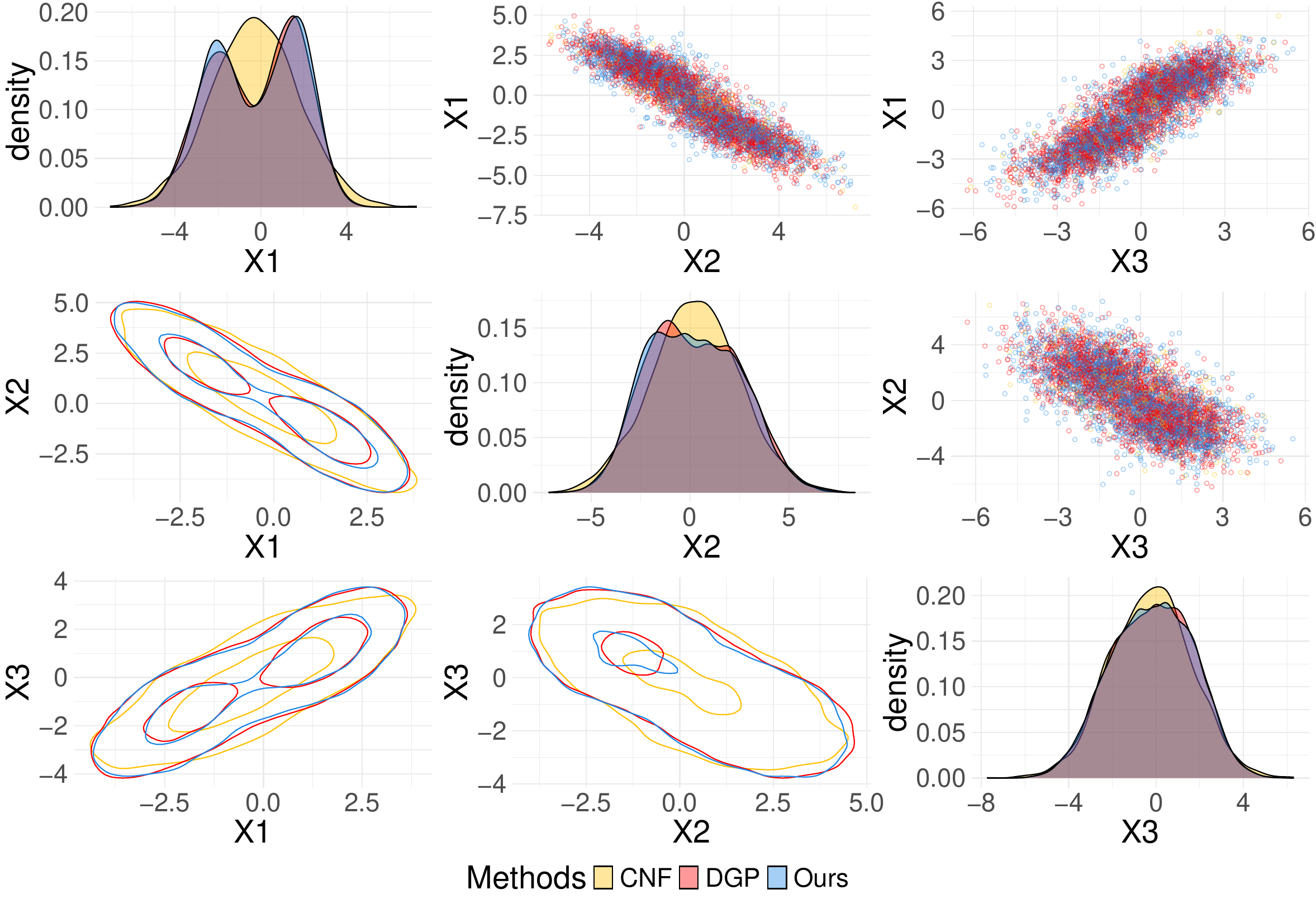} 
    \caption{$\mathcal{L}_1$: Comparative analysis of joint and marginal observational distributions for samples generated by the  DGP (see \cref{sec:appendix_VACA} red), our fitted TRAM-DAG (blue), and the fitted CNF from the original study (yellow) (\cite{javaloy2024causal}). The diagonal shows estimates of the marginal distributions, the lower triangle shows 2D density estimates, and the upper triangle presents scatter plots with subsampling.}
    \label{fig:VACA1Triangle_L1}
\end{figure}
To illustrate TRAM-DAG’s ability to fit complex observational continuous data flexibly, we replicate an example originally introduced in \cite{sanchez-martinVACADesigningVariational2022} and fitted with their VACA method. The data generating process (DGP) consists of three variables, $X_1, X_2, X_3$, details are provided in \cref{sec:appendix_VACA}. The variable $X_1$ follows a bimodal distribution, and $X_2$ and $X_3$ are linearly dependent on $X_1$. This leads to non-Gaussian marginal distributions also for $X_2$ and $X_3$ (see diagonal in \cref{fig:VACA1Triangle_L1}). 
We benchmarked flexible TRAM-DAG with
Causal Normalizing Flow (CNF) \cite{javaloy2024causal}.
\cref{fig:VACA1Triangle_L1} shows that the distribution of the samples from the fitted TRAM-DAG model closely resamples the distribution of the DGP, including the bimodal distribution of $X_1$, while the sample distribution from the fitted CNF fails to capture the bimodal distribution of $X_1$, likely due to the inflexibility of the used transformation.
We then adapted Neural Spline Flows (NSF) \citep{durkanNeuralSplineFlows2019} to the causal setting and observed a similar performance (see \cref{fig:VACA1Triangle_L1_NSF}) as achieved by TRAM-DAGs. This highlights the importance of achieving flexibility in distribution modeling,  which can be easily achieved with TRAM-DAGs.

\subsection{Interventional Distribution $\mathcal{L}_2$}
\label{sec:vaca_l2}
\begin{figure}[ht]
    \centering
    \includegraphics[width=0.6\columnwidth]{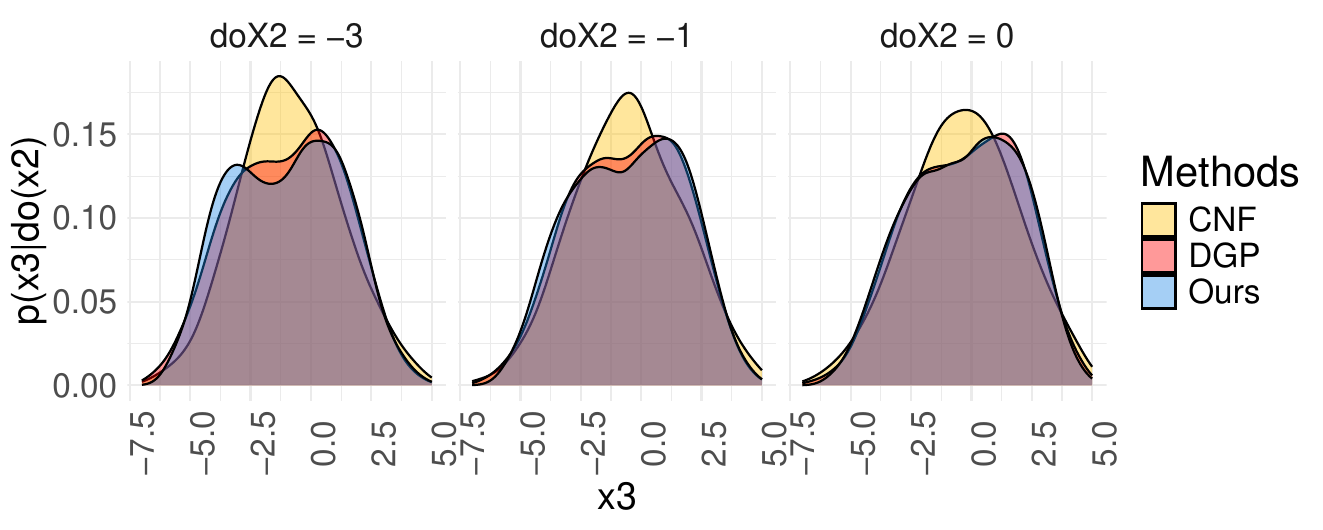} 
    \caption{$\mathcal{L}_2$: Interventional Distribution $P(X_3|\dd{X_2})$ for different values of the do-intervention resulting from the DGP (see \cref{sec:appendix_VACA} red), our fitted TRAM-DAG (blue), and the fitted CNF from the original study (yellow) (\cite{javaloy2024causal}). }
    \label{fig:VACA1Triangle_L2}
\end{figure}
Next, we benchmark TRAM-DAG’s capability to model interventional distributions by replicating the interventional experiment in \cite{javaloy2024causal}.   We used the fitted complex intercept TRAM-DAG to perform 
do-interventions on 
$X_2$ with $\dd{X_2}=-3$, $\dd{X_2}=-2$, and $\dd{X_2}=0$. Comparing the ground truth interventional distribution with the estimated interventional distribution achieved with the CNF method and our TRAM-DAG method \cref{fig:VACA1Triangle_L2}
demonstrates TRAM-DAG’s ability to capture the correct and complex interventional distribution and outperforms CNF. 
Please note that it is straightforward to estimate from the interventional distribution 
the treatment effect of increasing $X_2$ by one unit from $-3$ to $-2$ by the difference of the means of the estimated distributions, $\mathbb{E}(X_3|\dd{x_2=-2}) - \mathbb{E}(X_3|\dd{x_2=-3}$.

\subsection{Counterfactual queries $\mathcal{L}_3$} 
\label{sec:carefl}
To evaluate TRAM-DAG’s performance on counterfactual queries, we replicate the counterfactual experiments presented 
in CAREFL \citep{khemakhemCausalAutoregressiveFlows2021} based on a non-linear DGP with four variables (see \cref{sec:appendix_CAREFL} for more details).
%
Following \cite{khemakhemCausalAutoregressiveFlows2021}, we focus on the DGP generated observations, from which we pick the following observation
$x_{\text{obs}} = (2.00, 1.50, 0.81, -0.28)$. We then consider two counterfactual queries: 

(i) What would the expected value of $x_3$ would have been, if the variable $X_2$ would have taken the values  $x_2 = \alpha$ instead of observed value $x_2 = 1.5$? 

(ii) What would the expected value of $x_4$ would have been, if $X_1$ would have taken the value $x_1 = \alpha$ instead of the observed $x_1 = 2$? 

\begin{figure}[ht]
\centerline{\includegraphics[width=.8\columnwidth]{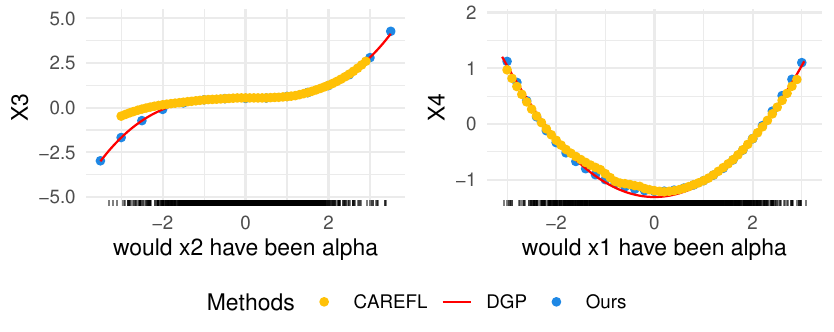}}
\caption{$\mathcal{L}_3$: Results of the counterfactual queries as posed in CAREFL based on a picked four-dimensional observation  \citep{khemakhemCausalAutoregressiveFlows2021}. 
\textbf{Left:} the expected value of the counterfactual distribution of $X_3$ (depicted on the y-axis) if $X_2$ would have taken the value $\alpha$ (depicted on the x-axis) instead of the observed value $x_2 = 1.5$.  
\textbf{Right:} the expected value of the counterfactual distribution of $X_4$ if $X_1$ would have been $\alpha$ instead of the observed value $x_1 = 2$. 
Shown are results from the DGP (see \cref{sec:appendix_CAREFL} red), our TRAM-DAG (blue) and CAREFL (yellow).
}
\label{fig:carefel_fig5}
\end{figure}

In both experiments $\alpha$ values in the range between $-3$ and $3$ are considered (see \cref{fig:carefel_fig5}).
The results of this benchmark experiment show that our TRAM-DAG closely resamples the true counterfactual results and slightly outperforms the CAREFL method.

Overall, TRAM-DAG is on par or slightly outperforms state-of-the-art NN- and NF-based causal methods on all three levels of Pearl's hierarchy. 

\section{Experiments with interpretable components} 

The following experiments focus on demonstrating that an SCM where the causal relationships are given by interpretable functions can be fitted by TRAM-DAGs without losing the interpretability. 
That is because \revision{causal} TRAMs can be set up as interpretable models \revision{allowing the user to understand and judge the modeled causal effect of each parent on the target variable, as well as predicting the effect of interventions (see \cref{sec:interpretation}, \cref{sec:a_interpretation_mixed})}. 

\subsection{Continuous Case} \label{subsec:cont}
\label{sec:inter_cont}
\begin{figure}[h]
\centering
\begin{minipage}{0.40\textwidth}
\centering
\includegraphics[width=\linewidth]{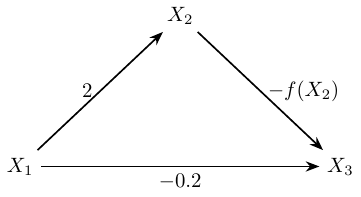}
\end{minipage}
\begin{minipage}{0.5\textwidth}
\centering
\includegraphics[width=0.8\linewidth]{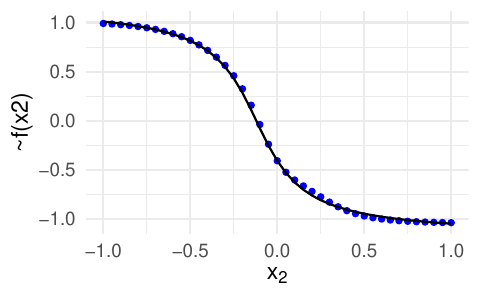} 
\hfill
\end{minipage}
\caption{\textbf{Left:} DAG with three continuous variables and meta information for shift terms in $h$: $h(x_1)=SI$, $h(x_2|x_1)=SI+2\cdot x_1$, $h(x_3|x_1,x_2)=SI-0.2\cdot x_1-f(x_2)$ \\
\textbf{Right:}  Complex shift term $CS=-f(x_2)$ depicted as solid black line for the  DGP ground truth $f(x_2)=0.75 \cdot \arctan\left(5 \cdot (x_2 + 0.12)\right)$ along with blue dots for the estimated CS in the fitted TRAM-DAG.}
\label{fig:DGP_cont}
\end{figure}

We start with an SCM involving three continuous variables (see DAG in \cref{fig:DGP_cont}). 

The DGP generates all three variables using a TRAM model with simple intercept and interpretable shift terms. 
We now perform three experiments, each with slightly different DGP and TRAM-DAG specifications. In the DGPs of all three experiments we use linear effects of $X_1$ on $X_2$ and $X_3$ ($\beta_{12}=2$, $\beta_{13}=-0.2$) and also use linear shift term to specify these effects in the fitted TRAM-DAGs.
However, we use different choices for the functional form $f$ of the causal effect of $X_2$ on $X_3$ in the DGP (see e.g.\ \cref{fig:DGP_cont}) and different specifications in the  fitted TRAM-DAGs.

\paragraph{Linear-shift DGP and linear-shift model}\, 
We  use in the DGP  $f(X_2) = -0.3 \cdot X_2$, resulting in in $h(x_3|x_1,x_2)=SI+\beta_{13}\cdot x_1-f(x_2)=SI-0.2\cdot x_1+0.3\cdot x_2$ and fit a correctly specified TRAM-DAG model with linear shift terms for all variables. The estimated coefficients   are in good agreement with the true values  from the DGP ($\beta_{12}=2$, $\hat{\beta}_{12}=1.98$ ; $\beta_{13}=-0.2$, $\hat{\beta}_{13}=-0.21$ ;  $\beta_{23}=0.3$, $\hat{\beta}_{23}=0.26$) \revision{and can be causally interpreted as log-odds-ratio (see \cref{sec:interpretation})}. The fitting process in shown in \cref{sec:appendix_inter_cont}.

\paragraph{Complex-shift DGP and complex-shift model}\,
Next, we increase the complexity of the DGP by defining $f(X_2)=0.75 \cdot \arctan\left(5 \cdot (X_2 + 0.12)\right)$ which introduces a non-linear causal impact of $X_2$ on $X_3$. We fit correctly specified TRAM-DAG with a complex shift term and achieve correctly estimated coefficients in the linear shift terms of $X_1$ on $X_2$ and $X_3$  ($\beta_{12}=2$, $\hat{\beta}_{12}=2.07$ ; $\beta_{13}=-0.2$, $\hat{\beta}_{13}=-0.203$ ) and a well fitted CS function $f(X_2)$ (see right panel in \cref{fig:DGP_cont}) 
 confirming TRAM-DAG’s capability to capture complex non-linear dependencies. 

\paragraph{Linear-shift DGP and complex-shift model}\,
Now we  use  in the DGP again a linear effect of $X_2$ on $X_3$ with $f(X_2) = -0.3 \cdot X_2$. However, we fit a miss-specified TRAM-DAG model with a complex shift term to model the effect of $X_2$ on $X_3$. 
Even under such a misspecification 
of the TRAM-DAG, the linear form of $f(x2)$ is approximately matched, and the coefficients $\beta_{12}=2$, $\beta_{13}=-0.2$ are well estimated (see Appendix \ref{sec:complex_model_simple_data}). Noteworthy, the observational and interventional distributions are accurately estimated (see \cref{fig:triangle_DPGLinear_ModelCS_L0_L1}).

Additional results of these experiments can be found in \cref{sec:appendix_inter_cont}.

\subsection{Mixed data types} \label{subsec:mixed}
\begin{figure}[ht]
\centering
\includegraphics[width=0.7\columnwidth]{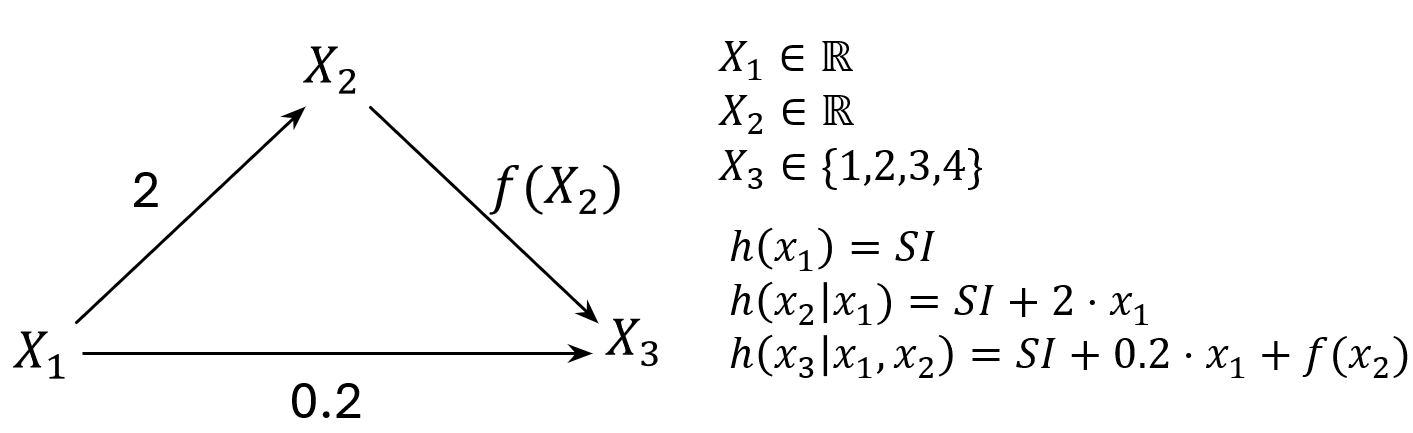}
\caption{DAG and setting for the interpretation experiment in the mixed case.}
\label{fig:dag_mixed_interpretation}.
\end{figure}

We now use an SCM involving three variables of mixed data types  (see \cref{fig:dag_mixed_interpretation}). Compared to the DGP from the continuous case in the last subsection we have replaced the continuous \( X_3 \) with an ordered categorical variable \( X_3 \in \{1,2,3,4\} \) (as in \cref{fig:trams}) and now use a positive linear effect of $X_1$ on $X_3$ ($\beta_{13}=2$) and a negative effect ($\beta_{23}=-0.3$) if $f$ is linear.

\begin{figure}[ht]
\centering
\includegraphics[width=0.75\columnwidth]{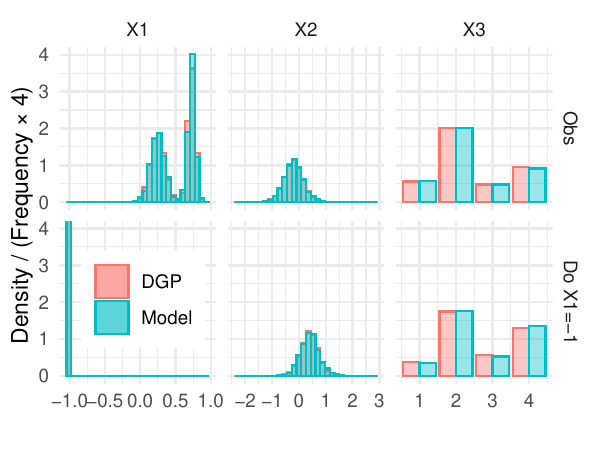}
\caption{$\mathcal{L}_1$ and $\mathcal{L}_2$ for the mixed data experiment: Comparison of observational and interventional distributions  where all shift therms were linear in the DGP and the fitted TRAM-DAG. The frequencies of $X_3$  have been multiplied by a factor of 4 for visual convenience.}
\label{fig:triangle_struct_mixed_L0L1}
\end{figure}

In a first experiment with mixed data types we used for $f(x_2)=-0.3\cdot x_2$. This results in $h(x_3|x_1,x_2)=SI+\beta_{13}\cdot x_1+f(x_2)=SI-0.2\cdot x_1-0.3\cdot x_2$  and showed that the coefficients in all three linear shift terms are accurately fitted (see \cref{fig:triangle_struct_mixed_coefs}). Then we showed that that observational and interventional distributions are accurately estimated in case of LS and CS (see (\cref{fig:triangle_struct_mixed_L0L1}, \cref{fig:triangle_struct_mixed_L0L1_exp}).


\revision{We have added an illustrative example in \cref{sec:a_interpretation_mixed} to demonstrated that the estimated coefficients in the linear shift terms of a correctly specified and fitted TRAM-DAG allow to correctly predict the effect of an intervention on a parent variable in terms of the resulting change of the odds to observe values below or equal to a freely specifiable cutoff after the intervention compared to before.  

With these experiments we have demonstrate TRAM-DAG's applicability to tackle $\mathcal{L}_1$ and $\mathcal{L}_2$ tasks with mixed data types.

\section{Conclusion}
In this paper, we introduced TRAM-DAGs, a novel framework for interpretable neural causal models. 
TRAM-DAGs range from transparent and interpretable causal models to causal models with the flexibility of deep learning models. 
Tuning the level of interpretability and flexibility for certain applications depends on the complexity of the data and the needed interpretability. 
Continuous TRAM-DAGs can be trained using observational data and used to answer queries across all three levels of Pearl's causal hierarchy: observational($\mathcal{L}_1$), interventional ($\mathcal{L}_2$), and counterfactual ($\mathcal{L}_3$). Mixed TRAM-DAGs are restricted to queries within the first two levels ($\mathcal{L}_1$ and $\mathcal{L}_2$). 
The possibility to incorporate binary, ordinal, continuous, or mixed data types in a TRAM-DAG is a big advantage compared to other state of the art causal models that rely on NNs.
In the continuous case, TRAM-DAGs fall within the class of bijective generation mechanism (BGM) models, inheriting all BGM properties, particularly their applicability to common causal structures with unobserved confounding \citep{nasr2023counterfactual}. 


\subsubsection*{Acknowledgements}  
We want to thank Lucas Kook for helpful discussions and Pascal Bühler for his help with the figures. We sincerely thank the reviewers for their valuable feedback.  
This work was partially supported by Carl-Zeiss-Stiftung in the project "DeepCarbPlanner" (grant no. P2021-08-007).

\clearpage

\bibliography{bibliography}

\appendix

\onecolumn

\section*{SUPPLEMENTARY MATERIAL}

\section{Transformation models}
\label{sec:app_trams}
TRAMs were introduced in 2014 as a flexible distributional regression method for tabular ordered data which can be continuous, discrete, or censored. \citep{hothorn2014conditional}. Later, TRAMs were extended to deep TRAMs by \cite{sick2021deep} using neural networks, allowing the inclusion of unstructured data modalities like images. 
TRAMs comprise most classical statistical regression models, like linear or logistic regression or other GLMs, and have hence the same interpretability of their parameters and the same guarantees as these well-established statistical models \citep{hothorn2014conditional}. However, TRAMs do provide a much larger family of models since TRAMs do not require pre-specify the family of the outcome distribution and allow to model flexible outcome distributions that change with the predictors, resulting in distributions that do not even need to belong to a known distribution family. 

\subsection{Interpretability of the shift terms}\label{sec:A_interpretability}

The choice of the latent distribution $F_u$ has no influence of the prediction power of the TRAM but determines the interpretation scale of the shift terms in the transformation function $h$ \citep{hothorn2014conditional}. In our experiments we  always use the standard logistic distribution $P(Y\le y) = F_Y(y)=F_{SL}(z):= (1 + \exp(-z))^{-1}$ as latent distribution \revision{with inverse $F_{SL}^{-1}(P)=\log\left(\frac{P}{1-P}\right)=\log(\odds)$} that allows to interpret the shift parameters as log-odds ratios. This is known from the logistic regression where the target $Y$ is binary, but is also valid for ordinal or continuous target variables as  demonstrated here for a $SI-LS_{x1}-CS_{x2}$ model with $h(y|x_1, x_2) = h_0(y) + \beta_1 x_1 + \gamma(x_2) $ and a continuous target $Y$:
%
\begin{align*}
F_{Y|x_1, x_2}(y) &= F_{SL}(h(y|x_1, x_2))
\\
\Leftrightarrow P(Y \leq y |x_1, x_2) &= 
\pSL(h_0(y) + \beta_1 x_1 + \gamma(x_2)) \\
\Leftrightarrow  \log(\odds(Y \leq y |x_1, x_2) &= 
h_0(y) + \beta_1 x_1 + \gamma(x_2) \\
\text{OR}_{x_1\rightarrow x_1+1} &=\frac{\odds(Y \leq y |x_1+1, x_2)}{\odds(Y \leq y |x_1, x_2)}
 =\frac{\exp{(h_0(y) + \beta_1 (x_1+1) + \gamma(x_2))}}{\exp{(h_0(y) + \beta_1 x_1 + \gamma(x_2))}} \\
\ &= \exp(\beta_1)
\\
\text{OR}_{x_2\rightarrow x_2+1} &= \frac{\odds(Y \leq y |x_1, x_2+1)}{\odds(Y \leq y |x_1, x_2)}
 =\frac{\exp{(h_0(y) + \beta_1 x_1 + \gamma(x_2+1))}}{\exp{(h_0(y) + \beta_1 x_1 + \gamma(x_2))}} \\
\ &= \exp(\gamma(x_2+1) - \gamma(x_2)) 
\end{align*}

Hence $\exp(\beta_i)$ is interpreted as odds-ratio, which is the factor by which the odds for $Y\leq y$ is changing if increasing $x_i$ by one unit and holding all other variables constant. Remarkably, this holds for any threshold value $y$, and hence, these models are called proportional odds models. 
Equivalently, the parameter $\beta_i$ in a  $LS$ term can be interpreted as log-odds-ratio
$\beta_i=\log(\text{OR}{x_i\rightarrow x_i+1})$.   \revision{Please note that in a causal model, where all predictors $X_i$ are direct causal parents of the target $Y$, this interpretation holds causally. This means that when intervening on the predictor $X_i$ by increasing it by one unit, the other parents of $Y$ may change upon this intervention and the observed odds of $Y\leq y$ in the interventional data will differ by the factor $\exp(\beta_i)$ compared to the observational data. We demonstrate that $\beta$ in a LS does correctly predict this interventional effect in an illustrative example in \cref{sec:a_interpretation_mixed}. This requires that the causal model is correct, meaning it matches the data-generating process
as in most of our experiments.
}

Proportional odds models are more commonly used for an ordinal target where the parameters in the LS terms quantify the change of the odds for $Y\leq y_k$ holding for all class levels $y_k$. 

The same math works for a binary target where we look at the odds for $Y\leq 0$, which is same as the odds for $Y = 0$. The odds for $Y = 0$  changes by the factor $\exp(\beta_i)$ if $x_i$ is increased by one unit and all other predictors stay constant. Note that in most implementations of the logistic regression the default latent distribution is the standard logistic distribution and a $\beta_i$ in the linear regression can be interpreted as log-odds-rations for $Y=1$.

For a $CS$ term, the difference of the estimated shifts can be interpreted as log-odds-ratio $\gamma(x_j+1) - \gamma(x_j) = \log(\text{OR}{x_j\rightarrow x_j+1})$ 

\section{Impossibility of Counter-Factual queries for discrete targets}
\label{sec:l3_discrete}

\revision{
Here we show why counterfactual queries can in general not be answered for discrete variables that are generated by censoring an underlying continuous variable. For example sport grades in an 100 meter running test are ordinal and give an incomplete quantification of the student's speed  since all students who run the 100 meter in a certain interval of time get the same grade - the grades are interval censored.

Counterfactual queries typically require a unique mapping from the observed outcome $\mathbf{x}$ to the underlying noise realizations $\mathbf{u}$ ("abduction"), so that one can subsequently "re-run" $\mathbf{u}$ under a hypothetical intervention ("action" and "prediction"). For continuous outcome variables, this mapping can be made bijective under mild assumptions, rendering counterfactual queries well-defined. However, for discrete (e.g., binary or ordinal) outcome variables, the mapping from a continuous noise variable to the observed discrete value is necessarily many-to-one: an entire interval of latent noise values collapses to a single discrete outcome.
}

\begin{figure}[h]
 \centering
    \includegraphics[width=.6
    \columnwidth]{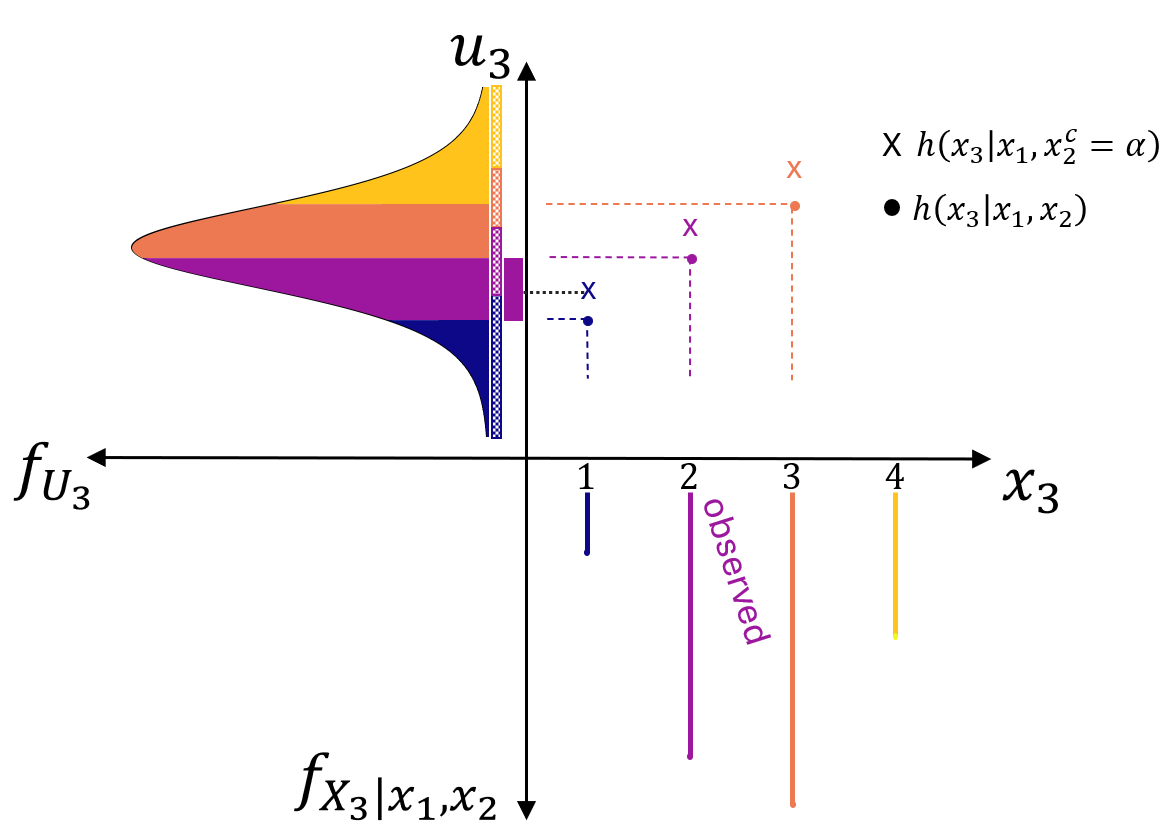}
\caption{TRAM with discrete target $X_3$ with a original $h$ corresponding to the observed values $x_1, x_2$ (indicated with dots) and a counterfactual $h$ in the counterfactual situation $x_2^c=\alpha$ (indicated as crosses). The \revision{solid} colored areas in $f_{U_3}$ correspond to the cutpoints of the original $h$, while the counterfactual $h$ would lead to shifted cutpoints as indicated by the \revision{hatched} colored bars below $f_{U_3}$. \revision{If $x_3=2$ was observed then the corresponding noise value could have been any value in the following interval $u_3 \in [h(1|x_1,x_2), h(2|x_1, x_2)]$ indicated by the thick purple bar. In the counterfactual situation where $X_2$ would have taken the value $\alpha$ the unambiguity of $u_3$ results in an unambiguity of the counterfactual value $x_3^c $, because different parts of the possible noise values (solid purple bar) fall into different counterfactual bins (hashed bars) which would result in different} counterfactual values of $x_3^c$.} 
\label{fig:discrete} 
\end{figure}

For illustration, look at Figures \ref{fig:trams} and \ref{fig:discrete}.
Let's  take the following counterfactual query: What value would $X_3$ have taken if the variable $X_2$ would have taken the values $\alpha$ instead of the observed value $x_2$?
For the discrete variable $X_3$, the discrete transformation $h(x_3|\pa{x_3})$  strictly monotone but not bijective. 
Hence, it is not possible to determine unambiguous values for the noise variable \revision{$U_3$} as needed in the abduction step of a counterfactual analysis . 
Imagine the observed level of $X_3$ was  level two, $x_3=2$ (purple), then all noise values $u_3 \in [h(1|x_1,x_2), h(2|x_1, x_2)]$ (indicated by the purple bar) would be possible since all these noise values lead to the observed value $x_3=2$. 
This can then lead to problems in the prediction step. In a counterfactual situation we imagine that $X_2$ would have taken $x_2^c=\alpha$ instead of the observed $x_2$. Hence, the transformation function $h(x_3|x_1, \alpha)$ (indicated by crosses in the Figure) has probably changed compared to the original $h(x_3|x_1, x_2)$ (indicated by dots in the Figure). Then it can happen that  $h(1|x_1, \alpha) \in [h(1|x_1,x_2), h(2|x_1, x_2)]$ (as illustrateted in the Figure), resulting in $x_3^c=1$, if 
$u_3 \leq h(1|x_1, \alpha)$  and $x_3^c=2$, if 
$h(1|x_1, \alpha) \leq u_3 \leq h(2|x_1, \alpha)$. As demonstrated in this example, it is not in general possible to determine an unambiguous counterfactual $x_3^c$ for ordinal variables.
\section{Additional Experimental Results and details of the experiments}
The code to reproduce the experiments is available at: \url{https://github.com/tensorchiefs/tram-dag} 

\subsection{Observational Distribution and do interventions}
\label{sec:appendix_VACA}
\subsubsection{Data Generating Process}
The Data Generating Process (DGP) follows the original VACA paper (see appendix E.1 in \cite{sanchez-martinVACADesigningVariational2022}). 
\begin{figure}[h] 
 \centering
    \includegraphics[width=.4\columnwidth]{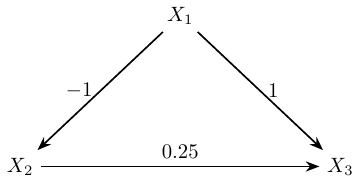}
\caption{DAG of the DGP process in the original VACA paper}
\label{fig:dag_vaca}
\end{figure}

$X_1 = \begin{cases} 
\mathcal{N}(-2, \sqrt{1.5})  \text{with probability } 0.5, \\
\mathcal{N}(1.5, 1)  \text{with probability } 0.5
\end{cases} $ 

$X_2 = -X_1 + \mathcal{N}(0, 1)$ 

$X_3 = X_1 + 0.25 \cdot X_2 + \mathcal{N}(0, 1)$ \\

Note that $X_1$  follows a bimodal distribution according to the following DGP

\subsubsection{Models}


%
We compare our TRAM-DAG against default implementation of  Causal Normalizing Flow (CNF) by  \cite{javaloy2024causal}. The CNF is based on MAF-like NN with 3 hidden layers each of dimension 16 using affine linear transformations but is still not able to fit a bimodal observational or interventional distribution (see \cref{fig:VACA1Triangle_L1}, \cref{fig:VACA1Triangle_L2}). 
It is important to note that \cite{javaloy2024causal} uses a different version of the DGP compared to \cite{sanchez-martinVACADesigningVariational2022}, where the bimodal distribution for $X_1$ is replaced by a standard normal distribution $\mathcal{N}(0,1)$ which could be fitted by CNF model.


\subsubsection{Additional Experiments with Neural Spline Flows}
In Figure \ref{fig:VACA1Triangle_L1_NSF}, we replicate the experiment from Figure \ref{fig:VACA1Triangle_L1}, but replace the inflexible Causal Normalizing Flow (CNF) presented by \cite{javaloy2024causal} with a Neural Spline Flow (NSF). As expected, the NSF achieves a more accurate fit to the bimodal distribution of $X_1$, underscoring the value of using flexible transformations, such as NSF or TRAM-DAGs with complex intercepts modeled by Bernstein polynomials for modeling complex distributions. 

\begin{figure}[ht]
    \centering
    \includegraphics[width=0.8\columnwidth]{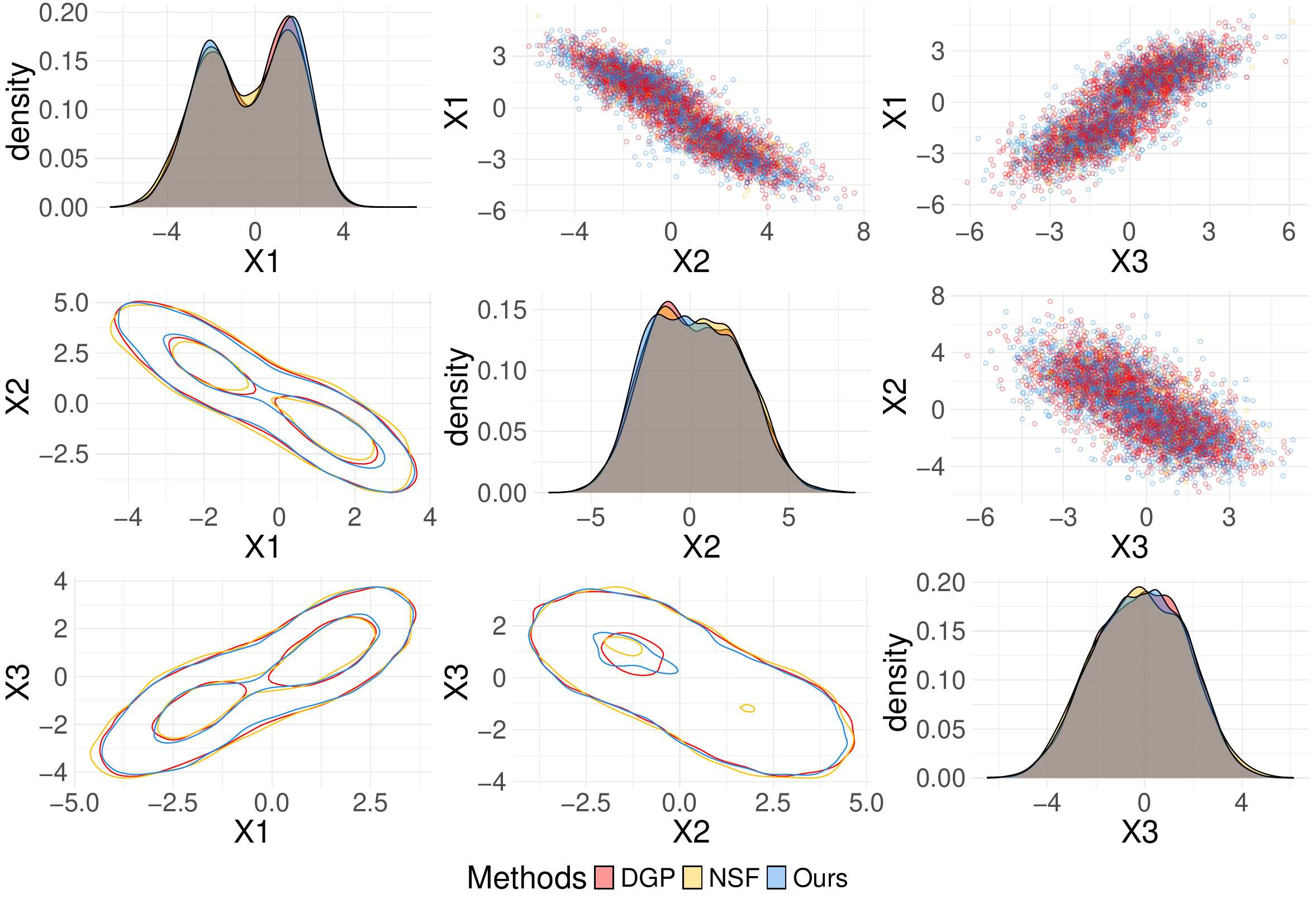} 
    \caption{$\mathcal{L}_1:$ Samples generated by the  DGP (see \cref{fig:dag_vaca}), our fitted TRAM-DAG and fitted Neural Spline Flow (NSF). Same experiment as for Figure~\ref{fig:VACA1Triangle_L1} but this time the inflexible CNF has been replaced by a flexible NSF.}
    \label{fig:VACA1Triangle_L1_NSF}
\end{figure}

\subsection{Counterfactual CAREFL experiment}
\label{sec:appendix_CAREFL}
Here we present additional 
information for the counterfactual CARELF experiment done by \cite{khemakhemCausalAutoregressiveFlows2021}.
The DAG 
is depicted in the following causal graph in \cref{fig:DAG_CAREFL}. 
\begin{figure}[h]
 \centering
    \includegraphics[width=.6\columnwidth]{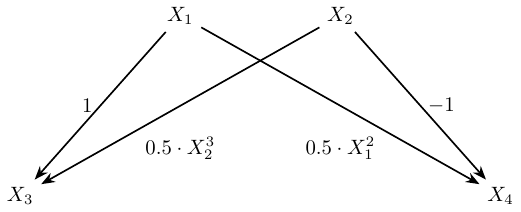}\
    \caption{DAG of the DGP used in the counterfactual experiment to benchmark TRAM-DAG with the CAREFL method presented in \ref{fig:DAG_CAREFL}}
    \label{fig:DAG_CAREFL}
\end{figure}
The DGP for the four variables $X_1$, $X_2$, $X_3$, and $X_4$, where $X_3$ and $X_4$ holds as non-linear transformations of $X_1$ and $X_2$ and is defined as follows:
\begin{align}
X_1, X_2 &\sim \text{Laplace}(0, \frac{1}{\sqrt{2}}) \nonumber \\
X_3 &= X_1 + 0.5 \cdot X_2^3 + \text{Laplace}(0, \frac{1}{\sqrt{2}}) \nonumber\\
X_4 &= -X_2 + 0.5 \cdot X_1^2 + \text{Laplace}(0, \frac{1}{\sqrt{2}})\nonumber
\end{align}

For our implementation, we employed the same architecture and training procedure as described in \cref{sec:MAF}. We compare our results against the original CAREFL model, as presented in Figure 5 of \cite{khemakhemCausalAutoregressiveFlows2021}.

\subsection{Interpretable Experiments Continuous Case}
\label{sec:appendix_inter_cont}
Here, we give additional results for the experiments with DGP with three continuous variables and fitted interpretable TRAM-DAGs (see \cref{subsec:cont}, \cref{fig:DGP_cont}). 

For the simple intercept in the interpretable continuous TRAM-DAG, we used Bernstein polynomials of order $M = 20$ (see \cref{eq:bernstein_h}). The training was conducted for 500 epochs on a dataset containing 40000 samples, utilizing the Adam optimizer with the default learning rate of 0.001.

\subsubsection{Linear-shift DGP and linear-shift model}\,
The causal effect of $X_2$ on $X_3$ is in the DGP is linear given by $f(X_2) = -0.3 \cdot X_2$, resulting in in $h(x_3|x_1,x_2)=SI+\beta_{13}\cdot x_1-f(x_2)=SI-0.2\cdot x_1+0.3\cdot x_2$ and hence $\beta_{23} = 0.3$. The used TRAM-DAG models the causal impact of all parents on their target as linear shift term. \cref{fig:triangle_struct_cont_coefs}  illustrates the evolution of the estimated coefficients throughout the training process, showing how the estimated coefficients in LS-terms of the TRAM-DAG converge towards  the true coefficients  of the DGP  $\beta_{12} = 2$,  $\beta_{13} = -0.2$ , and  $\beta_{23} = 0.3$. Please note, that the fitted TRAM-DAG accurately recovers the ground truth coefficients.
\begin{figure}[ht]
    \centering                             
    \includegraphics[width=0.9\columnwidth]{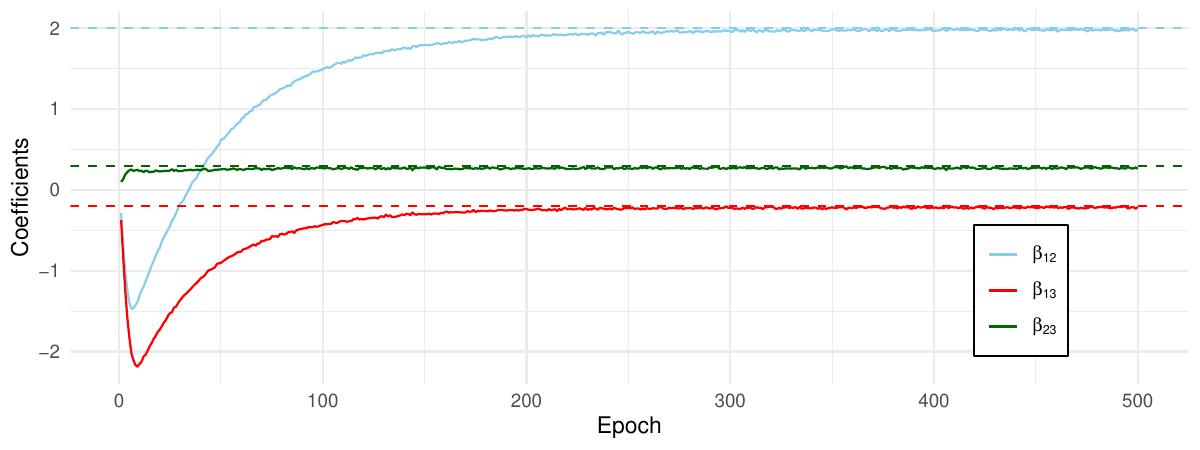} 
    \caption{Interpretable continuous case experiment with three linear shift terms: Estimated coefficients  $\beta_{12}$,  $\beta_{13}$, and  $\beta_{23}$  over training epochs, with dashed lines indicating the true coefficient values of the DGP.}
    \label{fig:triangle_struct_cont_coefs}
\end{figure}

\subsubsection{Complex-shift DGP and complex-shift model}\,
The causal effect of $X_2$ on $X_3$ in the DGP is modeled as complex shift$f(x_2)=0.75 \cdot \arctan\left(5 \cdot (x_2 + 0.12)\right)$. The other causal effects in the DGP remain linear effects.
 
Figure \cref{fig:triangle_mixed_DPGatan_ModelCS_coef_epoch} shows that the estimated coefficients in LS-terms of the TRAM-DAG converge towards  the true coefficients  of the DGP  $\beta_{12} = 2$,  $\beta_{13} = -0.2$.
\begin{figure}[ht]
    \centering
    \includegraphics[width=0.9\columnwidth]{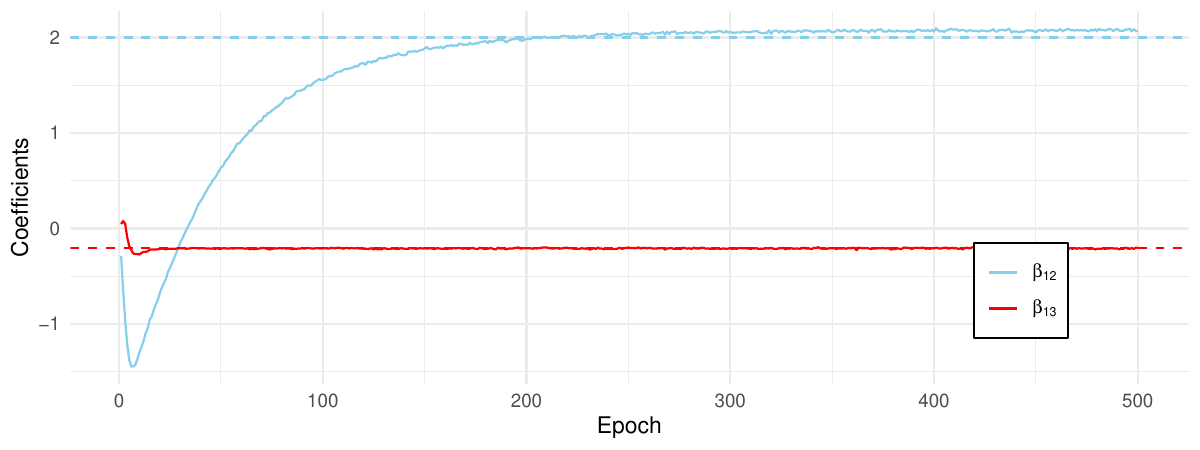} 
    \caption{Interpretable continuous case experiment with two linear shift terms and one complex shift term: Estimated coefficients  $\beta_{12}$,  $\beta_{13}$ of the two LS over training epochs, with dashed lines indicating the true coefficient values of the DGP. }
\label{fig:triangle_mixed_DPGatan_ModelCS_coef_epoch}
\end{figure}

In \cref{fig:triangle__DPGatan_ModelCS_L0_L1}, we see that the estimated observational and interventional distributions match the corresponding distributions produced by the DGP.
\begin{figure}[ht]
\centering
\includegraphics[width=0.75\columnwidth]{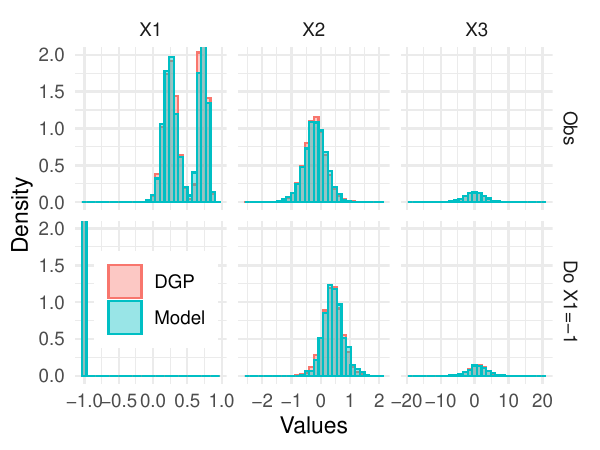}
\caption{$\mathcal{L}_1$ and $\mathcal{L}_2$ in the interpretable continuous case experiment with two linear shift terms and one complex shift term: Comparison of observational distributions (upper panel) and interventional distributions (lower panel) in the continuous case as generated by the DGP  or the fitted TRAM-DAG. The DGP holds $f(x_2)=0.75 * \text{atan}(5 * (x_2 + 0.12))$ and the TRAM-DAG modeled this effect by a CS.}
\label{fig:triangle__DPGatan_ModelCS_L0_L1}.
\end{figure}

\subsubsection{Linear Shift DGP, Complex Shift Model}
\label{sec:complex_model_simple_data}
We now examine the scenario of a misspecified TRAM-DAG is fitted. The causal effect of $X_2$ on $X_3$ is in the DGP is linear $\beta_{23}=0.3$, the other causal effects in the DGP are also linear effects with true coefficients   $\beta_{12} = 2$,  $\beta_{13} = -0.2$. However the TRAM-DAG models the causal effect of $X_2$ on $X_3$  by a complex shift. 
In \cref{fig:triangle_DPGLinear_ModelCS_L0_L1} (left side), we see the estimated complex shift term $-\hat{f}(X_2)$ for the case where the true function is $f(X_2) = -0.3 \cdot X_2$. We note small deviations of the fitted CS-term to the underlying linear function $f(x_2)$.
%
As seen in the right panel of \cref{fig:triangle_DPGLinear_ModelCS_L0_L1}, both the observational distributions and interventional distributions show close alignment between the DGP and our trained model. The minor derivations have little effect on the quality of the estimated observational distributions, making training more challenging. Therefore we attribute the slight deviations to the limited training time and time data.
\begin{figure}[ht]
    \centering
    \begin{minipage}{0.35\textwidth}
        \centering
        \includegraphics[width=\columnwidth]{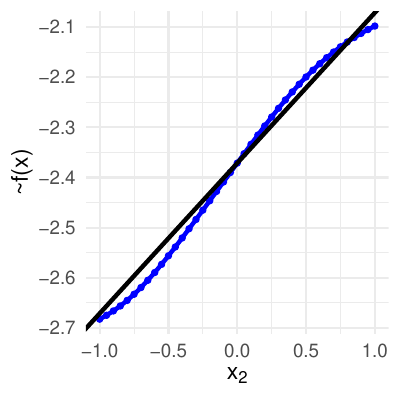}
    \end{minipage}
    \hfill
    \begin{minipage}{0.64\textwidth}
        \centering
        \includegraphics[width=\columnwidth]{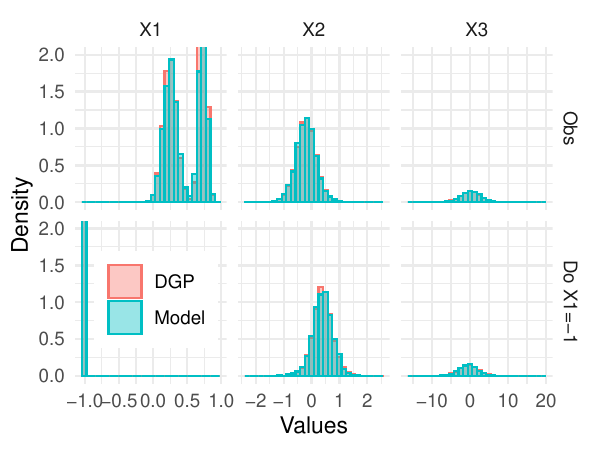}
    \end{minipage}
    \caption{Interpretable continuous case experiment with three linear shift in DGP and two LS and one CS in the model: \textbf{Left:} Comparing the true linear effect $f(x_2) = 0.3 \cdot x_2$ of $X_2$ on $X_3$ in the DGP (black solid line) with the estimated CS $\hat{f}(X_2)$ in the fitted TRAM-DAG (blue dots). \textbf{Right:} Comparison of observational distributions (upper panel) and interventional distributions (lower panel) as generated by the DGP or the fitted TRAM-DAG. }
    \label{fig:triangle_DPGLinear_ModelCS_L0_L1}
\end{figure}

\subsubsection{Non-monotonous DGP}
\revision{
To demonstrate that TRAM-DAGS are also able to fit complex non-monotonous functions the SCM, \cref{fig:triangle_DPGSinCS_L0_L1} presents the estimated coefficients $\beta_{12}$ and $\beta_{23}$, along with the estimated non-monotonous function $\hat{f}(x_2)$ for $f(x_2) = 2\sin(3 x_2) + x_2$.
}
\begin{figure}[ht]
    \centering
    \begin{minipage}{0.35\textwidth}
        \centering
        \includegraphics[width=\columnwidth]{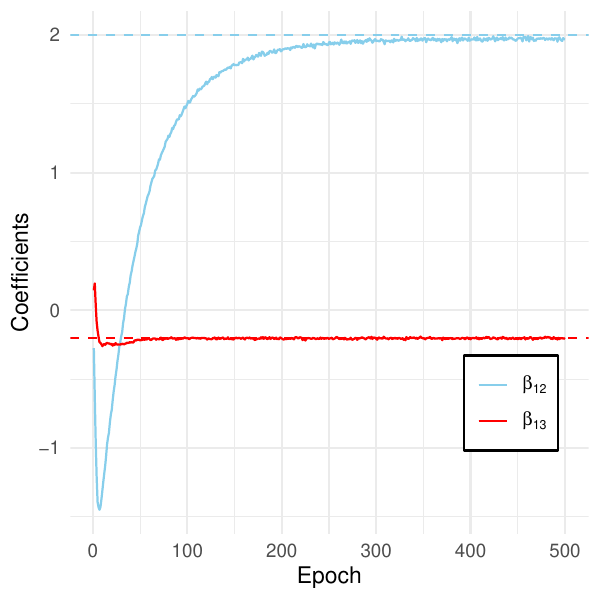}
    \end{minipage}
    \hfill
    \begin{minipage}{0.45\textwidth}
        \centering
        \includegraphics[width=\columnwidth]{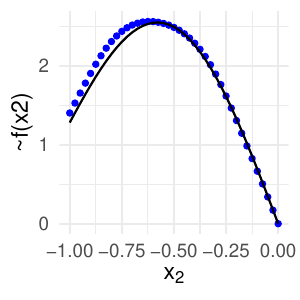}
    \end{minipage}
    \caption{
    Interpretable continuous case experiment with two linear shift terms and one non-monotone shift terms in DGP and a correctly specified model: \textbf{Left:} Comparison of the true and estimated coefficients in the linear shift terms. \textbf{Right:} Comparison of the true non-monotone shift term $f(x_2)=2 \sin(3 x_2)+x_2$ in the DGP (black solid line) and the fitted complex shift term in the TRAM-DAG (blue dots).}
    \label{fig:triangle_DPGSinCS_L0_L1}
\end{figure}

\clearpage

\subsection{Interpretable Experiments Mixed Case}
\label{sec:a_interpretation_mixed}
Here, we give additional results for the experiments for interpretable mixed TRAM-DAGs of \cref{subsec:mixed} where $X_3$ is an ordinal variable (see \cref{fig:dag_mixed_interpretation}). 


\paragraph{Linear-shift DGP and linear-shift model}\, 
We  use in the DGP  $f(X_2) = -0.3 \cdot X_2$ and fit a correctly specified TRAM-DAG model with linear shift terms for all variables. The estimated coefficients   are in good agreement with the true values  (see \cref{fig:triangle_struct_mixed_coefs})}..

\begin{figure}[h]
    \centering                             
    \includegraphics[width=0.8\columnwidth]
    {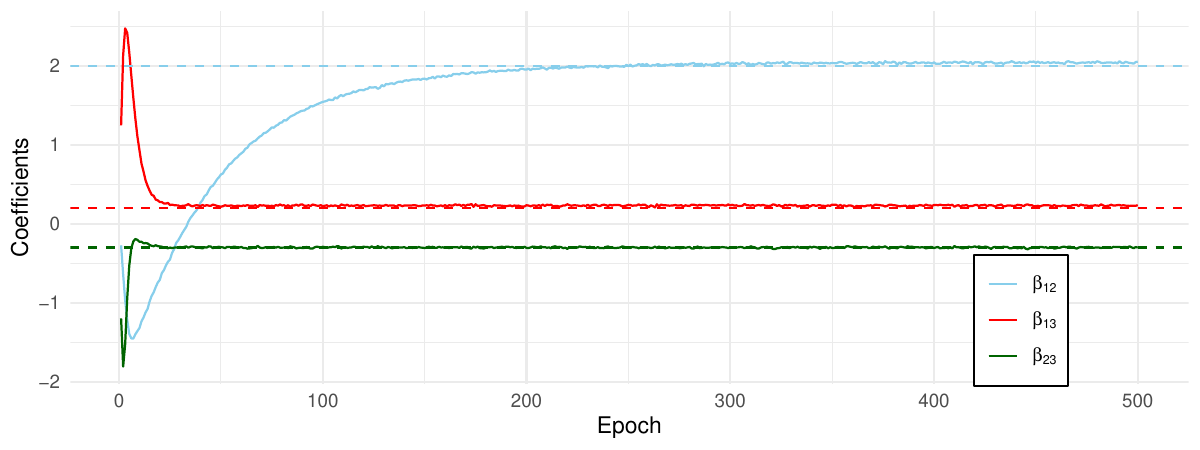} 
    \caption{Interpretable experiments mixed case with only linear shift terms: Estimated coefficients  $\beta_{12}$,  $\beta_{13}$, and  $\beta_{23}$  over training epochs, with dashed lines representing the true coefficient values of the DGP.}
    \label{fig:triangle_struct_mixed_coefs}
\end{figure}

\paragraph{Correctness of the predicted interventional effect}
To illustrate the interpretation of the  estimated causal coefficients in the linear shift terms we use the transformation function $h(x_2|x_1)=SI+\beta_{12}\cdot x_1$ where in the DGP $\beta_{12}=2$ which was estimated in the TRAM-DAG to be $\hat{\beta}_{12}=2.05$. 

Let's predict how the $\odds(x_2\leq c)=\frac{P(X_2\leq c)}{1-P(X_2\leq c)}$ will change if $x_1$ is increased by one unit and choose $c=-1$.

According the theory of causal TRAM-DAGs the $\odds(x_2\leq c)$ should be in the interventional data changed by the factor of $e^{\hat{\beta}_{12}}=e^{2.05}=7.74$ compared to this $\odds$ in the observational data. Hence the odds-ratio should be approximately $\hat{OR}=e^{2.05}=7.74$.

To check this prediction we sample from the original DGP $40000$ observations and then we adapt the DGP to a situation where $X_1$ is increased by one unit and sample also $40000$ observation from the interventional distribution. Using this data we count in both samples how many $x_2$ observations were greater or not then the arbitray chosen cutoff $c=-1$ and receive the following number:
\[
\begin{array}{lrr}
    \toprule
    & \multicolumn{2}{c}{\text{number of $x_2$-values}} \\
    \text{Type} & \leq -1 & > -1 \\
    \midrule
    \text{Interventional} & 5119  & 34881 \\
    \text{Observed}       &  744  & 39256 \\
    \bottomrule
\end{array}
\]
This leads a point estimate of $\hat{OR}=7.74$ and a 95\% confidence interval $[7.16, 8.38]$ that holds 
the prediction of $\hat{OR}=e^{\hat{\beta}_{12}}=e^{2.05}=7.74$ and the theoretical value of $OR=7.4$. 
Hence, we were able to predict the correct change of the $\odds(x_2\leq -1)$ upon increasing $X1$ in the DGP 
by one unit from the estimated parameter $\hat{\beta}_{12}=2.02$ 
of the TRAM-DAG that was fitted on observational data without having access to the interventional data.

\begin{figure}[ht]
\centering
\includegraphics[width=0.75\columnwidth]{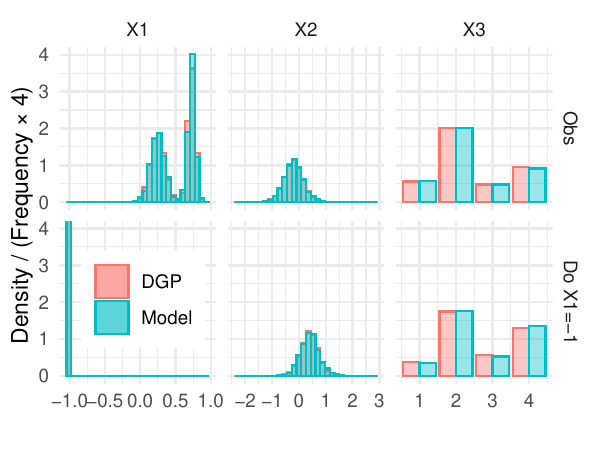}
\caption{$\mathcal{L}_1$ and $\mathcal{L}_2$ for the mixed data experiment  where the shift term from $X_2$ on $X_3$ were non-linear in the DGP with $f(x_2)=0.5 \cdot \exp(x_2)$ and modeled as  complex shift term in the fitted TRAM-DAG: Comparison of observational distributions (upper panel) and interventional distributions (lower panel) as generated by the mixed DGP or the fitted mixed TRAM-DAG. The frequencies of $X_3$  have been multiplied by a factor of 4 for visual convenience.  }
\label{fig:triangle_struct_mixed_L0L1_exp}.
\end{figure}

\paragraph{Complex-shift DGP and complex-shift model}\,
Next, we increase the complexity of the DGP by defining $f(x_2)=0.5 \cdot \exp(x_2)$ which introduces a non-linear causal impact of $X_2$ on $X_3$. We fit correctly specified TRAM-DAG and and show that the fitted model can be used to accurately estimate the observational and interventional distributions (see \cref{fig:triangle_struct_mixed_L0L1_exp}).

\end{document}